\documentclass[runningheads]{llncs}

\usepackage{accv}

\usepackage{accvabbrv}

\usepackage[accsupp]{axessibility}  

\usepackage{amsmath,amsfonts}
\usepackage{algorithmic}
\usepackage[ruled,vlined,linesnumbered]{algorithm2e}
\usepackage{array}

\usepackage{textcomp}
\usepackage{stfloats}
\usepackage{url}
\usepackage{verbatim}
\usepackage{graphicx}
\usepackage{cite}
\usepackage{comment}

\usepackage{tikz}
\usetikzlibrary{positioning}
\usepackage{pgfplots}
\pgfplotsset{compat=1.7}
\usepackage{diagbox}
\usepackage{comment}
\usepackage{amsmath}
\usepackage{booktabs} 
\usepackage{xspace}
\usepackage{comment}
\definecolor{commentcolor}{RGB}{110,154,155}   
\usepackage{color, colortbl}
\usepackage{bbding}
\usepackage{amsmath}

\usepackage{hyperref}

\usepackage{orcidlink}

\begin{document}

\title{DiDA: Video Object Segmentation with \underline{Di}stillation Learning of \underline{D}eformable \underline{A}ttention} 

\titlerunning{DiDA}

\author{Quang-Trung Truong\inst{1,2}\orcidlink{0000-0002-6242-2191} \and
Duc Thanh Nguyen\inst{3}\orcidlink{0000-0002-2285-2066} \and
Binh-Son Hua\inst{4}\orcidlink{0000-0002-5706-8634} \and \\
Sai-Kit Yeung\inst{1}\orcidlink{0000-0001-7974-0607}
}

\authorrunning{Truong et al.}

\institute{Hong Kong University of Science and Technology, Hong Kong SAR \and
Centre for Frontier AI Research, A*STAR, Singapore \and
Deakin University, Australia \and
Trinity College Dublin, Ireland}

\maketitle

\begin{abstract}

Video object segmentation is a fundamental research problem in computer vision. Recent techniques have often applied attention mechanism to object representation learning from video sequences. However, due to temporal changes in the video data, attention maps may not well align with the objects of interest across video frames, causing \textit{accumulated} errors in long-term video processing. In addition, existing techniques have utilised complex architectures, requiring \textit{highly} computational complexity and hence limiting the ability to integrate video object segmentation into low-powered devices. To address these issues, we propose \textbf{DiDA}, a new method for video object segmentation based on \textbf{Di}stillation Learning of \textbf{D}eformable \textbf{A}ttention. Specifically, we devise a lightweight architecture for video object segmentation that is effectively adapted to temporal changes. This is enabled by deformable attention mechanism, where the keys and values capturing the memory of a video sequence in the attention module have flexible locations updated across frames. The learnt object representations are thus adaptive to both the spatial and temporal dimensions. We train the proposed architecture using a new knowledge distillation paradigm where deformable attention maps are integrated into the \textit{distillation loss}. We qualitatively and quantitatively evaluate our method and compare it with existing methods on benchmark datasets including DAVIS 2016/2017 and YouTube-VOS 2018/2019. Experimental results verify the superiority of our method via its achieved state-of-the-art performance on YouTube-VOS18 dataset and optimal memory usage. Project page and code: \url{https://github.com/quangtrungtruong/DiDA}. 

\end{abstract}

\section{Introduction}
\label{sec:introduction}

\begin{figure}[ht]
\begin{subfigure}[t]{.55\linewidth}
\centering
\includegraphics[scale=0.4]{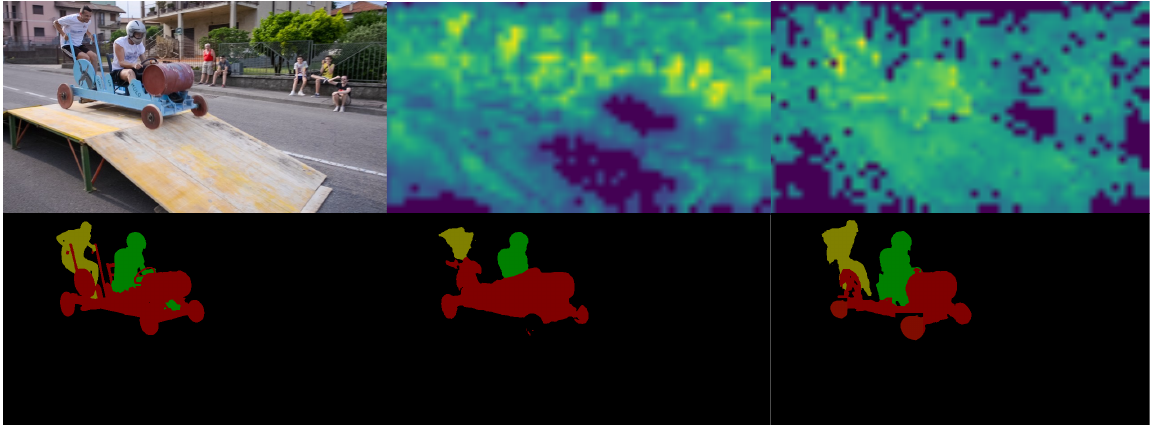}
\caption{}
\end{subfigure}\hfill%
\begin{subfigure}[t]{.42\linewidth}

\centering
\resizebox{0.7\textwidth}{!}{
\begin{tikzpicture}
\begin{axis}  
[  
    ybar,
    enlargelimits=0.4,
    legend style={
    cells={anchor=east},
    legend pos=north east,
    },
    label style={font=\Large},
    tick label style={font=\Large},
    xlabel={Number of query objects},
    ylabel={$\mathcal{J}$\&$\mathcal{F}$},
    symbolic x coords={1, 2, 3, 5},  
    grid=major,
    xtick=data,  
    nodes near coords,  
    nodes near coords align={vertical},  
    ]  
\addplot coordinates {(1, 90.05) (2, 70.55) (3,	71.7) (5, 80.7)}; 
\addplot coordinates {(1, 96.75) (2, 71.85) (3,	74.85) (5, 82.85)};  
\legend{MobileVOS~\cite{miles2023mobilevos}, Ours}  
\end{axis}  
\end{tikzpicture}}
\caption{}
\end{subfigure}
\caption{(a) From left to right: input frame and ground-truth segmentation masks, distilled feature map and segmentation masks by the distillation strategy used in MobileVOS~\cite{miles2023mobilevos}, distilled feature map and segmentation masks by our method.  (b) Segmentation accuracy ($\mathcal{J}$\&$\mathcal{F}$) of MobileVOS~\cite{miles2023mobilevos} and our method with different numbers of query objects in DAVIS-17 val (see more details in Section~\ref{sec:experiments}).}
\vspace{-0.3in}
\label{fig:teaser}
\end{figure}
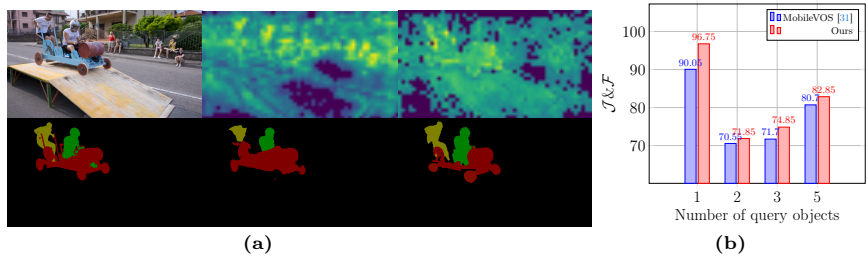

Video object segmentation (VOS) is a fundamental task in computer vision, aiming to segregate object(s) of interest from a background across frames in a video sequence. The task has attracted considerable attention from the research community, resulting in various models developed in recent years~\cite{DBLP:journals/air/GaoZYSDH23}. In the perspective of deep learning, designing an architecture that can well learn features of an object of interest adaptively to temporal changes while maintaining optimal memory usage is still an open research problem. Literature has shown a substantial body of work dedicated to developing deep learning models towards this goal~\cite{DBLP:journals/air/GaoZYSDH23}. Among these, the Vision Transformer (ViT) in~\cite{dosovitskiy2020image} has been commonly adopted in recent VOS research, and made significant progress. Examples include the works in~\cite{DBLP:conf/nips/YangWY21, DBLP:conf/cvpr/DukeA0AT21, DBLP:conf/nips/HeoHOLK22, Yoo2023, DBLP:conf/cvpr/ZhangLWXSZ23}. The reason for this success is the ability of the attention mechanism in the ViT in object representation learning. Specifically, unlike convolutional neural networks (CNNs) which obtain a global receptive field for an object by a pooling operator~\cite{richter2022receptive}, the ViT captures global context via self-attention layers. 

Despite such progress, there still exist issues in the current research. First, we found that attention layers are not always adapted to temporal change, causing \textit{accumulated errors} in processing of long-term video sequences. To mitigate this issue, several methods, e.g.,~\cite{tsai2016video, yu2022batman}, have utilised optical flows in the attention module and achieved promising results. Additional motion information from optical flows is a useful guideline to define an object in the query of the attention module. In particular, optical flows within the same object should be smooth while flows across the object boundaries should be disruptive. Similarly, if an object moves differently from the background, the motion boundaries would be indicative of the object boundaries. Hence, optical flows would facilitate precise locating of object boundaries and vice versa. However, these methods require an accurate motion estimation model to be given in advance. Unfortunately, this requirement is not always fulfilled, especially when VOS is applied to challenging scenarios such as underwater applications. 

Second, another major challenge in VOS is object forgetting in long-term video processing. The issue gets more critical in segmenting objects under severe occlusion. Several methods have been developed to tackle this challenge, e.g.,~\cite{DBLP:conf/cvpr/MiaoWY20, park2022per}. For instance, Park et al.~\cite{park2022per} indicated that memory updates in short-term intervals with several frames, also known as clip-wise mask propagation, are more powerful than updates with a nearby frame. However, one has to deal with clip-level optimisation and parallel computation of multiple frames. 

Third, existing methods are computationally expensive, limiting their applicability to low-powered devices. In particular, the computational complexity of ViT-based methods grows quadratically with their token length. The excessive number of keys to attend per query patch yields \textit{high} computational cost and \textit{slow} convergence, increasing the risk of over-fitting. Recently, MobileVOS~\cite{miles2023mobilevos} applied knowledge distillation to create a lightweight VOS model. The core idea of their distillation strategy is to constrain the similarity between distilled feature maps in a teacher and a student network via the similarity between their correlation matrices. However, we found such an approach is still affected by fast motion (under sudden changes due to fast motion, correlation matrices in distilled layers may be significantly diverse). We illustrate this issue in Fig.~\ref{fig:teaser}.

In this paper, we propose a VOS method to address these aforementioned issues. Specifically, we make the following contributions in our work.
\begin{itemize}
    \item We propose a deformable attention module for VOS to improve attention learning such that learnt attention maps are adaptive to both spatial and temporal changes. Our idea is motivated by Deformable Convolution Networks~\cite{dai2017deformable}, that learn a deformable receptive field for each convolution filter. We found such a learning approach is applicable to learning of attention maps and can be effective for driving attention scores to more informative regions, considering both the spatial and temporal dimensions. Here, we devise such a deformable attention-like pattern for VOS, where the positions of the key and value in the attention layer are not fixed but can be optimised from data. 
    
    \item We propose a lightweight architecture for VOS that can be trained using distillation mechanism. The learning process aims to transfer object representations learnt from a large model with full access to ground-truth labels to a smaller one with pseudo labels. We formulate this transfer learning process as knowledge distillation (KD). However, unlike existing KD methods which constrain only the consistency of the logits produced by the teacher and student networks, we further constrain intermediate attention maps in both the networks. 

    \item We prove the robustness of our method via extensive experiments on every aspect of its designs. In particular, we rigorously validate the core components including deformable attention, distillation learning of attention maps. We investigate various loss functions. We examine the distillation at different layers. We also compare our method with existing ones on benchmark datasets including YouTube-VOS 2019/2018 and DAVIS 2017/2016. Experimental results confirm the superiority of our method, showing its state-of-the-art performance on YouTube-VOS18 and optimal memory usage over the baselines. 
    
\end{itemize}

\section{Related work}

\label{sec:related_work}


\paragraph{Online-learning vs offline-learning.} Existing VOS methods can be categorised into online-learning methods or offline-learning methods, depending on how they train their model to segment a target object. Online-learning methods~\cite{caelles2017one, hu2018motion, perazzi2017learning} perform fine-tuning of a VOS model during the testing phase to incorporate specific information about the target object. Despite promising results, these methods are often experienced with over-fitting, i.e., they can learn the target object very well from the first frame, but fail to segment it in following frames. In addition, these methods are not practical for real-time applications as re-training of a model for a new object is time consuming. On the other hand, offline-learning methods~\cite{li2018video, cheng2021modular} aim to train a network that can work on any video without the need of re-training to adapt the model with a new object during testing. Our method follows the offline-learning approach, where we formulate VOS as label propagation over time.

\paragraph{Self-supervised learning.} This field receives the special attention of research community, recently. Self-supervised techniques utilise unlabeled videos with masks of the first frame given to identify query objects and then segment upcoming sequence frames in training. Generally, temporal correspondence learning is adopted to maintain temporal coherence in video segmentation methods such as Mining~\cite{jeon2021mining} and LIIR~\cite{li2022locality}. LIIR~\cite{li2022locality} belongs to temporal correspondence learning based on an additional video reconstruction. Inter-video and intra-video reconstruction scheme are used to formulate the contrast over inter-video and intra-video affinities to handle discriminating instances in the pixel-wise representation learning. Another family of temporal correspondence learning methods further execute both forward and backward tracking and penalize discrepancies between the initial and final positions of the considered pixels and regions, often known as cycle consistency based methods such as CorrFlow~\cite{Lai19} and Self-cycle~\cite{son2022contrastive}. Self-cycle~\cite{son2022contrastive} addresses noisy labels caused by the path selection problem in constructing a graph from a video. It benefits from the cycle-based temporal correspondences and hard negative mining in multi-hop concurrent path consideration. 

Another class of methods is mask correspondence learning. Recently, there are different mask-guided solutions aiming to generate a mask for self-supervised correspondence learning. Mask-VOS~\cite{li2023unified} creates pseudo-label data by adopting k-means clustering in order to enforce mask correspondence via a mask embedding scheme. Mask correspondence learning methods benefit from effective VOS methods belonging to semi-supervised learning that is trained under fully supervised fashion with real ground truth, e.g., DeAOTT~\cite{yang2022decoupling}. MAST~\cite{lai2020mast} shares the same with the matching-based method exploiting memory by calculating affinity matrix of query frames and reference frames. However, this framework is trained without manual annotation given. It is observed that the high-fidelity segmentation masks generated by semi-supervised VOS methods recently are effective to be exploited in self-supervised learning setting. The excellent matching-based architectures i.e., XMEM \cite{cheng2022xmem} or attention-based semi-VOS i.e., AOT \cite{DBLP:conf/nips/YangWY21}, DeAOT ~\cite{yang2022decoupling} enable to produce discriminative attention maps shown the superiority in modelling space-time correspondences. 
To the best of our knowledge, we are the first, coming up with a solution to exploit explicitly available powerful semi-supervised models in distillation mechanism. Specifically, mask correspondences are established between frames via attention paradigm DeAOT~\cite{yang2022decoupling}. Only pseudo labels are used in our proposed method with a simple but effective knowledge distillation framework to conduct attention transfer and logit transfer from a large teacher net to a student net for VOS.

\paragraph{Vision Transformers.} Vision Transformer (ViT), a network architecture inspired by the Transformer in~\cite{DBLP:conf/nips/VaswaniSPUJGKP17}, has shown its ability in various computer vision tasks, e.g., image recognition~\cite{dosovitskiy2020image}, semantic segmentation~\cite{Strudel_ICCV_2021}, and object detection~\cite{DBLP:conf/eccv/CarionMSUKZ20}. Such ability is enabled by attention mechanism~\cite{chen2020generative}, aiming to learn an attention map (score map) for every representation (e.g., a local image region) within a given context (e.g., an entire image). However, many areas in an attention map of a large-sized input may be dismissed during the training. To address this issue, several methods apply sliding window partition to the input data~\cite{ainslie2020etc, liu2021swin, dong2022cswin}. Dense attention in ViT is beneficial for learning of large receptive fields, but also incurs expensive memory usage and computational cost. To overcome this challenge, Xia et al.~\cite{xia2022vision} proposed deformable attention, where the offsets of the keys and values in the self-attention module are not fixed in a regular grid, but determined from data. Similarly, Pan et al.~\cite{pan2023slide} proposed slide-transformer, which allows location shifting of the key and value offsets. To shift the keys and values accordingly with depth-wise convolutions, the authors replaced the original column-based view to calculate the key and value matrices by a row-based view. There are methods combining both CNNs and ViTs. For instance, Xiao et al.~\cite{xiao2021early} applied convolutions in early stages of a ViT to enhance the stability of the model training. CSwin Transformer proposed in~\cite{dong2022cswin} employed convolution-based positional encoding and demonstrated significant improvement. These convolution-based techniques have the potential to be applied in conjunction with deformable attention to further enhance performance. In this paper, we devise a deformable attention-like pattern for ViT-based VOS. In sharp contrast to our proposed architecture, \cite{xia2022vision} proposed a large deformable attention transformer architecture in which the number of parameters of its variants vary between 48M and 69M parameters. Specifically, the model \cite{xia2022vision} consists of four stages of attention for image segmentation and detection tasks. Multiple-head deformable attention block is placed in the two last stages and the first two stages contain local attention and shift-window attention. On the contrary, our single-head deformable attention is designed right after the encoder block. Our light-weight attention student is effective since attention distillation is leveraged to mimic teacher's discriminative attention maps and transfer to student net in data-driven manner.

\paragraph{Knowledge distillation.} Knowledge distillation (KD) is a powerful machine learning technique that aims to transfer knowledge learnt from a large-sized model (teacher) to a smaller-sized one (student). 
For instance, Cheng et al.~\cite{cheng2023boxteacher} adopted KD for instance segmentation where only box-level labels are available for training. Many recent KD frameworks ~\cite{huang2022knowledge,chen2022improved, zong2022better, park2021learning} have focused on design of loss functions or adapters. MobileVOS~\cite{miles2023mobilevos} proposed a distillation loss based on pixel-wise multiplication of correlation matrices of distilled feature maps. Unlike existing methods, in this paper, we propose a KD scheme for VOS training, where the knowledge transfer between the teacher and student architectures happens not only in logit layers but also in attention maps.

\section{Proposed method}
\label{sec:proposed_method}

\subsection{Overview}
\label{sec:overview}

\begin{figure*}[!t]
\begin{center}
\centering
\includegraphics[width=1\textwidth]{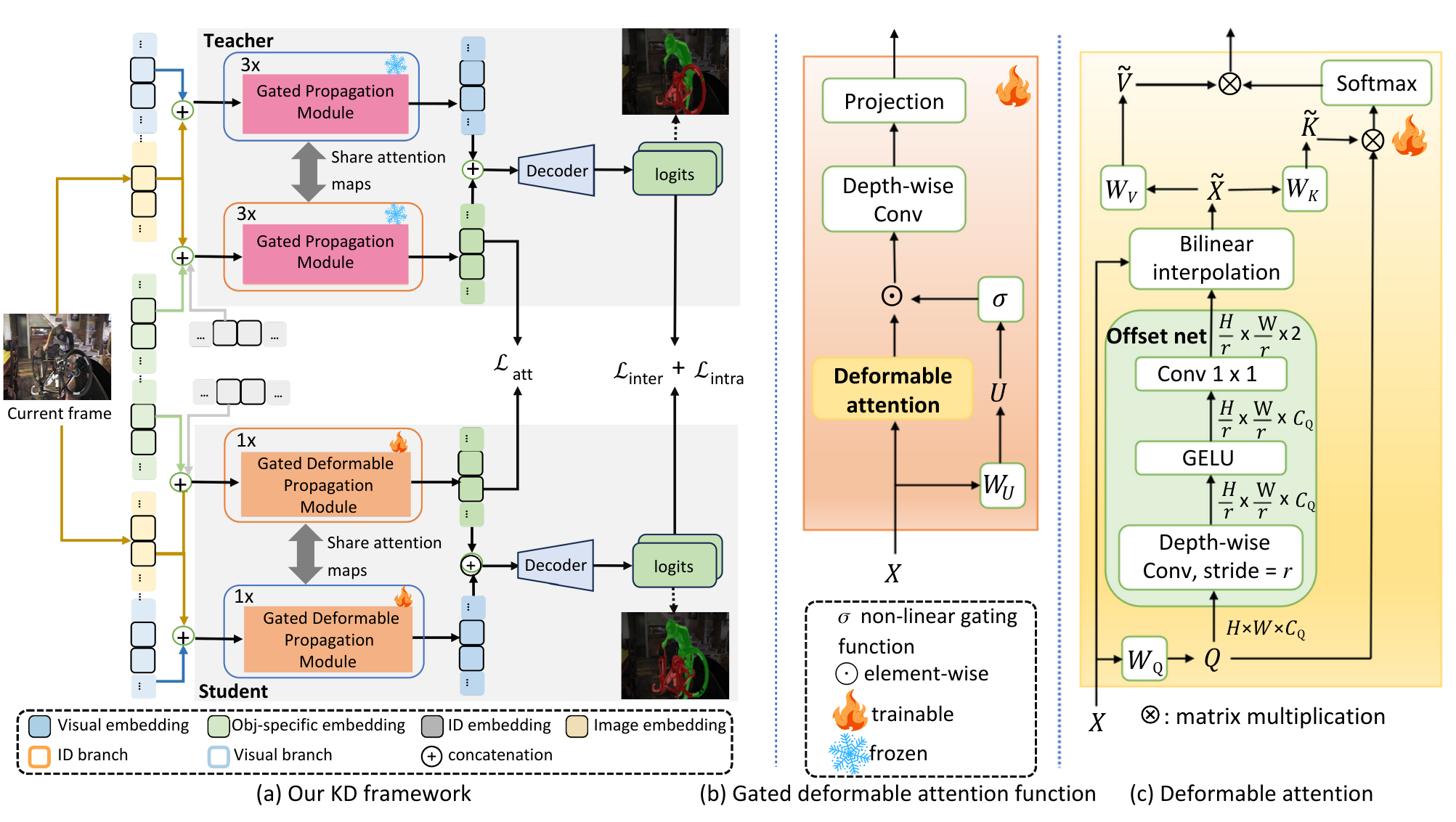}
\end{center}
   \caption{\textbf{Summary of our proposed VOS method}. (a) Overview of our knowledge distillation method. The teacher model transfers intermediate attention maps to the student model. This transfer is enforced by a CKA-based loss $\mathcal{L}_{att}$. At the same time, probability distributions of logits are transferred using intra-object and inter-object losses $\mathcal{L}_{intra}$ and $\mathcal{L}_{inter}$. Both the teacher and student models make use of Gated Propagation Module (GPM)~\cite{yang2022decoupling}, aiming to propagate spatio-temporal information across frames via the attention mechanism. (b) Our proposed Gated Deformable Attention function, which is used to implement the GPM. (c) Deformable attention module, which is used to replace the vanilla attention in the Gated deformable attention function.}
\label{fig:framework}
\end{figure*}

Our method aims at performing effective knowledge distillation (KD) for video object segmentation (VOS). We examine our method with DeAOT, the state-of-the-art VOS in~\cite{yang2022decoupling}. Specifically, we opt DeAOTL as our teacher network as this is the largest model among all the variants of the DeAOT's family. In addition, we build our student network upon DeAOTT, the smallest model in this group. An overview of our knowledge distillation framework is shown in Fig.~\ref{fig:framework}(a).

Both the teacher and student networks learn attention maps to share between two network branches: visual branch and ID branch. The visual branch aims to match objects by passing embeddings stored in memory across adjacent frames. The ID branch propagates object-specific knowledge learnt from past frames to the current frame to associate objects of the same ID across frames. In the teacher model, the shared attention maps are learnt by the Gated Propagation Module (GPM)~\cite{yang2022decoupling}. We refer the readers to our supplementary material for the implementation details of the GPM. To improve the adaptivity of attention maps to temporal changes, we propose to replace the Gated Attention function, used in the GPM, by our Gated Deformable Attention function (see Fig.~\ref{fig:framework}(b)) which is implemented via deformable attention (see Fig.~\ref{fig:framework}(c)). We present the deformable attention in Section~\ref{sec:deformable_attention}.

We apply KD in training of our VOS framework. We opt to distill the attention map and logit layer in the 3rd GPM of the ID branch from the teacher network to the student one. We describe our KD scheme in Section~\ref{sec:knowledge_distillation}. Unlike existing KD methods which transfer logit layers from the teacher to student model, here we also transfer attention maps during the training phase. In particular, the teacher model first transfers intermediate attention maps to the student model. These attention maps are calculated using our proposed deformable attention module. We constrain the attention map transfer via a Centered Kernel Alignment (CKA)-based loss~\cite{kornblith2019similarity}. Probability distributions of the logits (i.e., output of the softmax layer) are then transferred via intra-object and inter-object losses.

\subsection{Deformable attention module}
\label{sec:deformable_attention}

\subsubsection{Vanilla self-attention}

Since we focus on image-like data formats, for the convenience in presentation, we describe the vanilla self-attention for tensor-based inputs, e.g., a high-dimensional feature map $X \in \mathbb{R}^{H \times W \times C}$ where $H \times W$ represent the spatial dimensions and $C$ represents the number of channels. A more general definition can be found from~\cite{DBLP:conf/nips/VaswaniSPUJGKP17}.

Given a feature map $X$, the query $Q$, key $K$, and value $V$ in a single-head attention module are calculated as,
\begin{align}
    Q = W_q X, K = W_k X, V = W_v X
    \label{eq:basic}
\end{align}
where $W_q$, $W_k$, and $W_v$ contain learnable parameters.

The single-head self-attention transforms each query by calculating a weighted sum of values. The weights are computed by taking the dot product between the query and its corresponding keys, followed by a normalisation step as,
\begin{align}
    \text{Att}(Q, K, V) = \text{softmax}\left(\frac{Q{K^{\top}}}{\sqrt{d}}\right)V
    \label{eq:self_attention}
\end{align}
where $\sqrt{d}$ is a scaling factor in the attention mechanism. 

\subsubsection{Deformable attention}

Inspired by Deformable Convolution Networks~\cite{dai2017deformable}, deformable attention~\cite{xia2022vision} allows the offsets of the keys and values in the self-attention mechanism flexible yet learnable from data. Particularly, deformable attention calculates an attention map for an input feature map in 3 steps: 1) initialise reference points (locations for the keys and values), 2) generate offsets, 3) re-sample the features regarding to new reference points shifted the generated offsets. 

\textbf{Initialise reference points.} Deformable attention uses a set of irregular points, called reference points, to locate the query and key in the given feature map $X$. Those reference points are initialised from a uniform grid $R=\{(0,0),..,(H_g-1, W_g-1)\}$ where $H_g = H / g$ and $W_g = W / g$, $g$ is a grid-size factor. The reference points are then normalised into $[-1, 1]$. 

\textbf{Generate offsets.} The query $Q$ is calculated as in~\cref{eq:self_attention} and then partitioned evenly along its feature channels. In particular, let $Q \in \mathbb{R}^{H \times W \times C_Q}$, $Q$ is partitioned into $S$ sub-feature maps $\left \{ Q_i \in  \mathbb{R}^{H \times W \times C_{Q_i}} \right \}_{i=1}^S$, where $\sum_i^S C_{Q_i} = C_Q$ and $C_{Q_i} = C_{Q_j}$, $\forall i, j \in \{1,...,S\}$.

Each sub-feature map $Q_i$ is passed to an offset network $M(Q_i)$ to generate a set of offsets $\Delta_i = \{\delta_{i,r} \in \mathbb{R}^2 | \forall r \in R\} \in \mathbb{R}^{2 \times H_g \times W_g}$. The offset network is a convolutional neural network consisting of 2 convolutional layers with GELU activation functions in between (see our supplementary material for the details). Finally, given $Q$, we can calculate a set of offsets $\Delta = \left \{ \Delta_i\right \}_{i=1}^S \in \mathbb{R}^{2 \times S \times H_g \times W_g}$.

\textbf{Re-sample the features.} We re-sample the features in $X$ at new locations made by shifting the reference points with their offsets in $\Delta$. In particular, let $\Tilde{X}_{i}$ be the re-sampled feature map corresponding to the offsets $\Delta_i$. Let $\mathbf{p}_r \in \mathbb{R}^2$ be a location relative to a reference point $r \in R$. We calculate $\Tilde{X}_{i}(\mathbf{p}_r)$ using bilinear interpolation as follows,
\begin{align}
    \Tilde{X}_{i}(\mathbf{p}_r) = \sum_{\mathbf{q} \in R} I(\mathbf{p}_r + \delta_{i,r}, \mathbf{q}) X(\mathbf{q})
    \label{eq:resample}
\end{align}
where $I$ is defined as,
\begin{align}
    I(\mathbf{p}, \mathbf{q}) = &\max(0, 1-|\mathbf{p}_x-\mathbf{q}_x|) \times  \nonumber \\
    &\max(0, 1-|\mathbf{p}_y-\mathbf{q}_y|) 
    \label{eq:bilinear}
\end{align}
where $\mathbf{p}=(\mathbf{p}_x,\mathbf{p}_y) \in \mathbb{R}^2$ and $\mathbf{q}=(\mathbf{q}_x,\mathbf{q}_y) \in \mathbb{R}^2$. 

Intuitively, ~\cref{eq:resample} re-samples $\Tilde{X}_{i}(\mathbf{p}_r)$ based on the four discrete locations closest to $\mathbf{p}_r + \delta_{i,r}$. Next, we calculate deformable key $\Tilde{K}$ and deformable value $\Tilde{V}$ as,
\begin{align}
    \Tilde{K} = W_k \Tilde{X}, \Tilde{V} = W_v \Tilde{X}
    \label{eq:deformable_key_value}
\end{align}

Finally, a deformable attention map is achieved as,
\begin{align}
    \text{DefAtt}(Q, \Tilde{K}, \Tilde{V}) = \text{softmax} \left( \frac{Q{\Tilde{K}^\top}}{\sqrt{d}} \right) \Tilde{V}
    \label{eq:deformable_attention}
\end{align}

To prioritise important tokens in a video sequence, the deformable attention map is element-wise multiplied with a gating embedding as,
\begin{align}
    \text{GatedDefAtt}(Q, \Tilde{K}, \Tilde{V}, U) = \text{DefAtt}(Q, \Tilde{K}, \Tilde{V})\odot \sigma(U)
    \label{eq:gating_deformable_attention}
\end{align}
where $U = W_u X \in \mathbb{R}^{W \times H \times C_u}$ with $W_u$ as a learnable parameter matrix, $\sigma(\cdot)$ is a non-linear gating function (e.g., SiLU/Swish ~\cite{xia2022vision, ramachandran2017searching}), and $\odot$ is an element-wise product.

\subsection{Knowledge distillation}
\label{sec:knowledge_distillation}

Although we expect the student model to be smaller in size and faster at inference, we also aim to make it strong in terms of performance. To achieve this, we propose a new knowledge distillation scheme that guides the student network to learn not only logits from the teacher network but also intermediate attention maps. In addition, to further strengthen the distillation process, we enforce relational matching between the predictions of the student and the teacher network. In particular, we apply the intra-object and inter-object relations in~\cite{huang2022knowledge} in our distillation loss. These object-based relations fit well our purpose (object segmentation) and are proven to strengthen the prediction ability of our method against fast motion and deformable shapes. 

Let $F_T^k$ and $F_S^l$ be attention maps in the teacher and student networks at encoder block $k$ and $l$ respectively. In our implementation, we distill the last attention map in the teacher network. We enforce the consistency between $F_T^k$ and $F_S^l$ during the distillation process. Feature distillation has often been performed via projectors~\cite{chen2022improved, liu2023function}, where KL-divergence is utilised to reduce the discrepancy between corresponding layers in both the teacher and student models. However, we found that the KL-divergence loss is usually anisotropic. To effectively transfer attention maps between the teacher and the student networks, we define our loss based on the Centered Kernel Alignment (CKA)~\cite{kornblith2019similarity}, which is proven isotropic with respect to all dimensions regardless of scales, and reliant only on the feature distributions. Here we summarise the main steps of CKA calculation and refer the readers to the work by~\cite{kornblith2019similarity} for more details. First, we apply a linear kernel to both $F_T^k$ and $F_S^l$ to obtain Gram matrices $G_T^k$ and $G_S^l$. Let $H(G_T^k, G_S^l)$ be the Hilbert-Schmidt Independence Criterion-based metric~\cite{gretton2005measuring} between the Gram matrices $G_T^k$ and $G_S^l$. The CKA score between $F_T^k$ and $F_S^l$ is defined as,
\begin{equation}
    \text{CKA}(F_T^k,F_S^l) = \frac{H(G_T^k, G_S^l)}{\sqrt{H(G_T^k,G_T^l) \cdot H(G_S^k,G_S^l)}}
\end{equation}

The loss for attention distillation is finally defined as,
\begin{align}
    \mathcal{L}_{att} = \sum_{k} \sum_{l} a_{k,l} (1 - \text{CKA}(F_T^k,F_S^l))
    \label{eq:att_loss}
\end{align}
where $a_{k,l}=1$ if $F_T^k$ is transferred to $F_S^l$, and $a_{k,l}=0$, otherwise.

Next, we present how to distill logit values. Let $\mathbf{Z}_S \in \mathbb{R}^{B \times N}$ and $\mathbf{Z}_T \in \mathbb{R}^{B \times N}$ be the logit values from the student and teacher model on a batch of $B$ samples and $N$ objects. Let $\mathbf{Y}_S \in \mathbb{R}^{B \times N}$ and $\mathbf{Y}_T \in \mathbb{R}^{B \times N}$ be the probability distributions over the $N$ objects achieved from $\mathbf{Z}_S$ and $\mathbf{Z}_T$ using the softmax operator, i.e., $\mathbf{Y}_{S/T}[i,:]=\text{softmax}(\mathbf{Y}_{S/T}[i,:])$, $i=1,...,B$. The distillation loss for the inter-object and intra-object relations are defined as,
\begin{align}
\mathcal{L}_{inter} &= \frac{1}{B}\sum^B_{i=1}d_p(\mathbf{Y}_S[i,:], \mathbf{Y}_T[i,:]) \\
\mathcal{L}_{intra} &= \frac{1}{N}\sum^N_{j=1}d_p(\mathbf{Y}_S[:,j], \mathbf{Y}_T[:,j])
\label{eq:logit_loss}
\end{align}
where $d_p$ is the Pearson's distance measuring the discorrelation between two probability distributions.

The final loss of our distillation method is calculated as,
\begin{equation}
    \mathcal{L} = \mathcal{L}_{inter}+\mathcal{L}_{intra}+\lambda\mathcal{L}_{att}
    \label{eq:dist_loss}
\end{equation}
where $\lambda$ is a balancing factor. 

\section{Experiments}
\label{sec:experiments}

\subsection{Datasets} 
We conducted our experiments on VOS benchmark datasets including DAVIS 2016~\cite{perazzi2016benchmark}, DAVIS 2017~\cite{pont20172017}, YouTube-VOS 2018 and 2019~\cite{xu2018youtube}. The DAVIS 2016 consists of video sequences with single objects of interest. This dataset has 30 videos for training and 20 videos for validation with high-quality ground truth segmentation for salient objects. The DAVIS 2017 is an improved version of the DAVIS 2016 with 60 videos for training and 30 videos for validation. 

The YouTube-VOS~\cite{xu2018youtube} is a large-scale dataset for segmenting multiple objects. It has 3,471 videos for training, 474 and 507 videos for validation in the 2018 and 2019 version, respectively. The training set has 65 categories, and the validation set further includes 26 unseen categories. 

\subsection{Experimental setup}
\label{sec:experimental_setup}

\paragraph{Implementation details} 

We adopted the pre-trained DeAOTL model from~\cite{yang2022decoupling} with ResNet50 backbone as the teacher network. We chose the DeAOTT also from~\cite{yang2022decoupling} with MobileNet-V2 backbone as the student network. We set $\lambda=1.5$ in~\cref{eq:dist_loss}. 

We performed the distillation mechanism, i.e., no access to the ground-truth labels during the training of the student model. Specifically, we applied the teacher model to generate pseudo labels that are used to train the student model. Following the setting in~\cite{yang2022decoupling}, we first performed the distillation on synthetic video sequences generated from 28,732 images from BIG~\cite{cheng2020cascadepsp}, DUTS~\cite{wang2017learning}, ECSSD~\cite{shi2015hierarchical}, FSS-1000~\cite{li2020fss}, HRSOD~\cite{zeng2019towards} datasets. We then trained the student model on the VOS datasets (DAVIS~\cite{perazzi2016benchmark, pont20172017}, YouTube-VOS~\cite{xu2018youtube}). Data augmentation was applied to both the training steps. We conducted all experiments on two NVIDIA RTX-3090 GPUs. The training process on the synthetic and VOS datasets took 24 hours with 100K iterations and 18 hours with 130K iterations, respectively. We set the training batch size to 16 in the whole training process. 


\paragraph{Evaluation metrics.} We evaluated the performance of our method and other baselines using the standard VOS metrics including the region similarity score $\mathcal{J}$, the boundary accuracy $\mathcal{F}$, and their mean ($\mathcal{J} \& \mathcal{F})$. The $\mathcal{J}$ score is the average intersection-over-union ratio between predicted and the ground-truth masks. The $\mathcal{F}$ score is the average similarity between the boundary of predicted and the ground-truth masks). We also followed the evaluation protocol by \cite{perazzi2016benchmark} to calculate these metrics.

\begin{table*}[!t]
\centering
\caption{Comparison of the vanilla attention and deformable attention in attention learning in VOS. Best performances are highlighted.}
\scalebox{0.96}{
\begin{tabular}{l cccccccc}
\toprule
\rowcolor{Apricot} Attention mechanism & \multicolumn{3}{c}{DAVIS-16 Val}                                  & \multicolumn{3}{c}{DAVIS-17 Val}                                  & YT-VOS18 & YT-VOS19 \\ 
\cline{2-4}\cline{5-7}
\rowcolor{Apricot} & \multicolumn{1}{c}{$\mathcal{J}$\&$\mathcal{F} \uparrow$}   & \multicolumn{1}{c}{$\mathcal{J} \uparrow$}    & $\mathcal{F} \uparrow$    & \multicolumn{1}{c}{$\mathcal{J}$\&$\mathcal{F} \uparrow$}   & \multicolumn{1}{c}{$\mathcal{J} \uparrow$}    & $\mathcal{F} \uparrow$    & $\mathcal{J}$\&$\mathcal{F} \uparrow$           & $\mathcal{J}$\&$\mathcal{F} \uparrow$           \\ 
\midrule
Vanilla attention  & \multicolumn{1}{c}{83.35}  & \multicolumn{1}{c}{83.30}   & 83.40   & \multicolumn{1}{c}{69.10} & \multicolumn{1}{c}{66.60} & 71.60 & 68.61  & 69.26  \\ \hline
Deformable attention  & \multicolumn{1}{c}{\textbf{85.75}}     & \multicolumn{1}{c}{\textbf{84.90}}    &  \textbf{86.60}   & \multicolumn{1}{c}{\textbf{72.75}}  & \multicolumn{1}{c}{\textbf{69.90}} & \textbf{75.60} & \textbf{73.18}             & \textbf{73.95}          \\
\bottomrule
\end{tabular}}
\label{table:deformable_att}
\end{table*}

\begin{table*}[!t]
\centering
\caption{Comparison of state-of-the-art knowledge distillation methods. Best performances are highlighted. For~\cite{chen2022improved}, we re-executed the supplied code from the original work, but were not able to produce reasonable results on the Davis-16 and YT-VOS18/19 datasets. We, therefore, fill the results of~\cite{chen2022improved} on those datasets with ``-''.}
\begin{tabular}{l cccccccc}
\toprule
\rowcolor{Apricot}Methods & \multicolumn{3}{c}{DAVIS-16 Val}                                  & \multicolumn{3}{c}{DAVIS-17 Val}                                  & YT-VOS18 & YT-VOS19 \\ 
\cline{2-4}\cline{5-7}
\rowcolor{Apricot} & \multicolumn{1}{c}{$\mathcal{J}$\&$\mathcal{F} \uparrow$}   & \multicolumn{1}{c}{$\mathcal{J} \uparrow$}    & $\mathcal{F} \uparrow$    & \multicolumn{1}{c}{$\mathcal{J}$\&$\mathcal{F} \uparrow$}   & \multicolumn{1}{c}{$\mathcal{J} \uparrow$}    & $\mathcal{F} \uparrow$    & $\mathcal{J}$\&$\mathcal{F} \uparrow$           & $\mathcal{J}$\&$\mathcal{F} \uparrow$           \\
\midrule
DIST~\cite{huang2022knowledge}  & \multicolumn{1}{c}{84.50}  & \multicolumn{1}{c}{84.00}   & 85.00   & \multicolumn{1}{c}{71.05} & \multicolumn{1}{c}{68.50} & 73.60 & 71.40  & 72.90  \\ \hline
PEFD~\cite{chen2022improved}   & \multicolumn{1}{c}{-}     & \multicolumn{1}{c}{-}    & -    & \multicolumn{1}{c}{57.20}  & \multicolumn{1}{c}{54.80} & 59.60 & -             & -             \\ \hline
MobileVOS~\cite{miles2023mobilevos}   & \multicolumn{1}{c}{80.1}     & \multicolumn{1}{c}{79.5}    & 80.7    & \multicolumn{1}{c}{70.3}  & \multicolumn{1}{c}{67.9} & 72.7 & 72.68             & 72.96             \\ \hline
Ours   & \multicolumn{1}{c}{\textbf{85.75}} & \multicolumn{1}{c}{\textbf{84.90}} & \textbf{86.60} & \multicolumn{1}{c}{\textbf{72.75}} & \multicolumn{1}{c}{\textbf{69.90}} & \textbf{75.60} & \textbf{73.18}  & \textbf{73.95}  \\ 
\bottomrule
\end{tabular}
\label{table:kd}
\end{table*}

\begin{figure*}[t]
    \centering
    \includegraphics[width=1.0\textwidth]{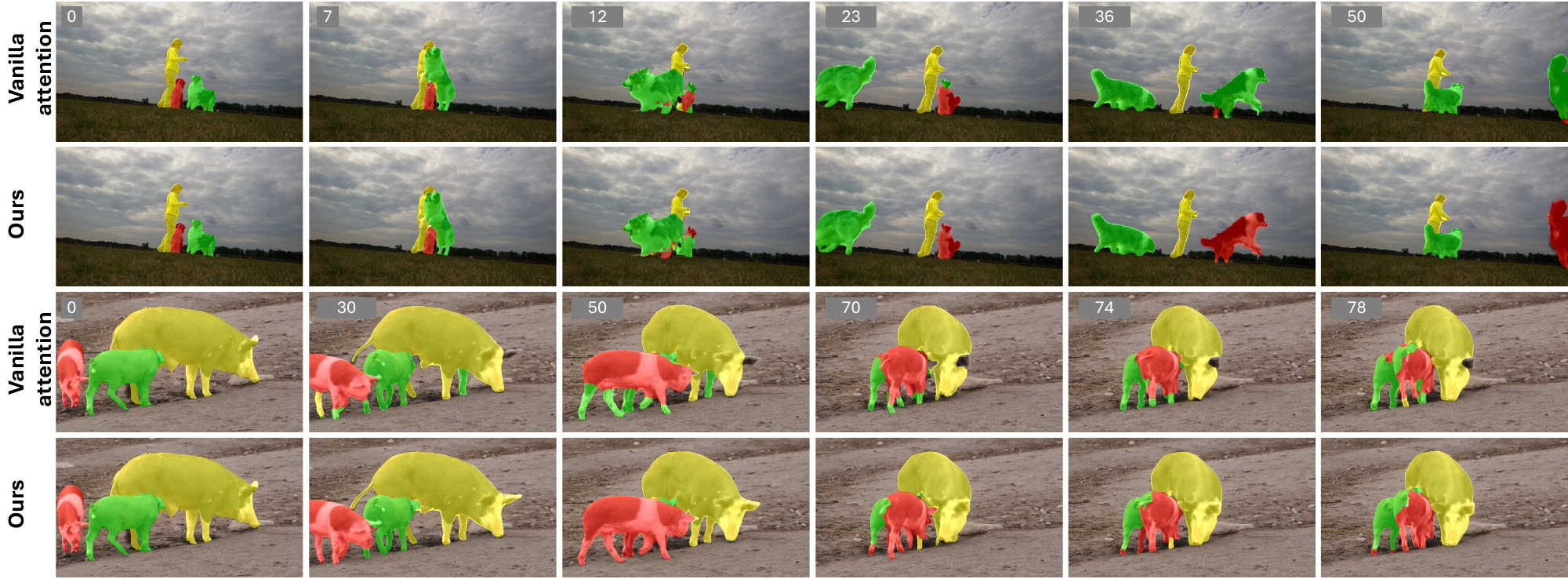}
    \caption{Qualitative results of our method and a baseline (a DeAOTT model with the vanilla attention trained using the standard KD, i.e., only logit layers are transferred). As shown, compared with the baseline, our method can maintain the association of the objects and their IDs across frames (see frame 36 in the 1st and 2nd row). Our method also tends to be aware of parts of the same object (see frame 50 in the 3rd and 4th row). 
    We hypothesize this success is due to the cross-frame adaptivity of deformable attention over its counterpart. More results are provided in our supplementary material.}
    \label{fig:samples}
\end{figure*}

\begin{table}[t]
\centering
\caption{Comparison of VOS methods in terms of segmentation accuracy ($\mathcal{J}\&\mathcal{F}$, $\mathcal{J}$, $\mathcal{F}$) and inference speed (frame-per-second - FPS) on DAVIS-17 Val dataset. Best and second-best performances are highlighted with bold and underlines, respectively. Works in~\cite{son2022contrastive,jeon2021mining} do not provide executable code for re-production. Their inference speed, thus, is not reported. Methods in~\cite{Lai19,son2022contrastive}, on the other hand, are not evaluated on the YouTube-VOS18 dataset. Note that we use pseudo-labels to train the student network, while the remaining models are trained in a self-supervised manner.}
\begin{tabular}{l cccc c}
\toprule
\rowcolor{Apricot}Methods & \multicolumn{4}{c}{DAVIS-17 Val}                                  & YouTube-VOS18   \\ 
 \cline{2-5}
\rowcolor{Apricot}                                     & \multicolumn{1}{c}{$\mathcal{J}$\&$\mathcal{F} \uparrow$}   & \multicolumn{1}{c}{$\mathcal{J} \uparrow$}    & $\mathcal{F} \uparrow$    &  FPS $\uparrow$     & $\mathcal{J}$\&$\mathcal{F} \uparrow$      \\
\midrule
\multicolumn{1}{l}{CorrFlow~\cite{Lai19}}           &  \multicolumn{1}{c}{50.30}  & \multicolumn{1}{c}{48.40} & 52.20 &       2.00 & -    \\ \hline
\multicolumn{1}{l}{MAST~\cite{lai2020mast}}              & \multicolumn{1}{c}{65.50}  & \multicolumn{1}{c}{63.30} & 67.60  & 2.06& 64.20              \\ \hline
\multicolumn{1}{l}{Self-cycle~\cite{son2022contrastive}} &  \multicolumn{1}{c}{70.50}  & \multicolumn{1}{c}{67.40} & 73.60 & -& -                  \\ \hline
\multicolumn{1}{l}{Mining~\cite{jeon2021mining}}         &  \multicolumn{1}{c}{70.30}  & \multicolumn{1}{c}{67.90} & 72.60 & - & 67.30              \\ \hline
\multicolumn{1}{l}{LIIR~\cite{li2022locality}}           &  \multicolumn{1}{c}{72.10}  & \multicolumn{1}{c}{69.70} & 74.50 & 1.87 & 69.30               \\ \hline
\multicolumn{1}{l}{Mask-VOS~\cite{li2023unified}}     &  \multicolumn{1}{c}{\textbf{74.50}}  & \multicolumn{1}{c}{\textbf{71.60}} & \textbf{77.40}     & 1.77 & \underline{71.60}       \\ \hline
\multicolumn{1}{l}{DeAOTT~\cite{yang2022decoupling}}     &  \multicolumn{1}{c}{69.10}  & \multicolumn{1}{c}{66.60} & 71.60  &  \textbf{64.26}& 68.61          \\ \hline
\multicolumn{1}{l}{Ours}                                 &  \multicolumn{1}{c}{\underline{72.75}} & \multicolumn{1}{c}{\underline{69.90}} & \underline{75.60}     & \underline{52.36}   & \textbf{73.18}      \\ 
\bottomrule
\end{tabular}
\label{table:self-supervised}
\end{table}

\subsection{Evaluation and results}

We first evaluate our proposed deformable attention in learning of attention maps in VOS. To do this, we experiment our framework (in Fig.~\ref{fig:framework}(a)) with the vanilla attention and deformable attention. We report the results of this experiment in Table~\ref{table:deformable_att}. As shown in the results, the deformable attention outperforms the vanilla one on all the evaluation metrics and across all the datasets. 

Next, we evaluate the effectiveness of our proposed KD method. Recall that our method also distills the attention maps, in addition to the logits during the distillation process. Therefore, to validate this idea, we compare our KD strategy with the one in~\cite{huang2022knowledge}, which transfers only the logits from the teacher to the student model. This strategy is equivalent to setting $\lambda$ (in~\cref{eq:dist_loss}) to 0. In addition, we experiment with the state-of-the-art KD algorithm in~\cite{chen2022improved}. For a fair comparison, we replicate the methods in~\cite{huang2022knowledge,chen2022improved,miles2023mobilevos} for the VOS scenario using the same experimental setup presented in Section~\ref{sec:experimental_setup}. We summarise the results of this comparison in Table~\ref{table:kd}. The experimental results confirm the benefit of distillation of intermediate attention maps during the distillation process. The results also show the superiority of our proposed KD method over the state-of-the-art KD. 

We compare our method with existing self-supervised VOS methods on common datasets and present the comparison results in Table~\ref{table:self-supervised}. In addition to evaluating the methods using segmentation accuracy metrics, we also compare their computational speed using frame-per-second (FPS) metric. As shown in Table~\ref{table:self-supervised}, our method ranks first on the YouTube-VOS 2018 dataset and second on the DAVIS 2017 in terms of the segmentation accuracy (evident by the $\mathcal{J}\&\mathcal{F}$ scores). However, compared with the first-ranked method, i.e., Mask-VOS~\cite{li2023unified}, although our method incurs a lower accuracy ($< 2.5\%$ of the $\mathcal{J}\&\mathcal{F}$ score), it takes much less memory due to the lightweight architecture yet offers a much faster inference speed (about 30 $\times$ of the FPS), making the VOS real-time and potential suitability for resource-constrained hardware. We summarise the segmentation accuracy vs memory footprint of all the methods in Fig.~\ref{fig:accuracy_vs_memory}. We visualise several qualitative results of our method in Fig.~\ref{fig:samples}.

\begin{figure}[ht]
\begin{minipage}[t]{0.495\textwidth}
\centering
\begin{tikzpicture}
\begin{axis}[legend pos=south east, 
    ymajorgrids=true,
    width=1.0\textwidth,
    grid=major,
    xlabel= memory (GB),ylabel=$\mathcal{J}$\&$\mathcal{F}$]
    \addplot[
        scatter,only marks,scatter src=explicit symbolic,
        scatter/classes={
            a={mark=pentagon*,blue},
            b={mark=triangle*,red},
            c={mark=o,draw=black,fill=black},
            e={mark=diamond*,green},
            f={mark=square*,red}
        }
    ]
    table[x=memory,y=avg,meta=label]{
        memory    avg    label
        8.86	66.5	a
        1.42	50.3	b
        16.95	72.1	c
        1.01    68.10   e
        1.26	72.75	f
    };
    \legend{MAST~\cite{lai2020mast},CorrFlow~\cite{Lai19},LIIR~\cite{li2022locality}, DeAOTT~\cite{yang2022decoupling}, Ours}
\end{axis}
\end{tikzpicture}
\captionsetup{font=footnotesize}
\captionof{figure}{Comparison of video object segmentation methods in terms of segmentation accuracy ($\mathcal{J}\&\mathcal{F}$) and memory footprint on DAVIS-17 Val dataset.}
\label{fig:accuracy_vs_memory}

\end{minipage}\hfill
\begin{minipage}[t]{0.48\textwidth}

\centering
\resizebox{1.0\textwidth}{!}{
\begin{tikzpicture}
\begin{axis}[
scaled y ticks=real:150,
ytick scale label code/.code={},
ymax = 4,
height=8cm,
width=8cm,
grid=major,
xlabel={iterations},
ylabel={loss score},
label style={font=\Large}, 
tick label style={font=\Large},
legend style={
cells={anchor=east},
legend pos=north east,
},
every axis plot/.append style={line width=2pt} 
]
\addplot table[x expr=\coordindex,y=s]{metric.txt};
\addplot table[x expr=\coordindex,y=logit]{metric.txt};
\addplot table[x expr=\coordindex,y=CKA]{metric.txt};
\addplot table[x expr=\coordindex,y=KL]{metric.txt};
\legend{Distillation loss ($\mathcal{L}$), Logit loss ($\mathcal{L}_{inter} + \mathcal{L}_{intra}$), Attention loss ($\mathcal{L}_{att}$) - CKA, Attention loss ($\mathcal{L}_{att}$) - KL}
\end{axis}
\end{tikzpicture}}
\captionsetup{font=footnotesize}
\captionof{figure}{Convergence analysis of the loss functions.}
\label{fig:losses}
\end{minipage}
\end{figure}

\subsection{Ablation study}

We conducted ablative experiments to explain the rationale behind the design and settings of our method. 

\begin{table*}[t]
\small
\begin{minipage}[t]{.41\linewidth}
\centering
\caption{Comparison of KL-divergence and CKA. Best performances are highlighted.}
\resizebox{0.96\textwidth}{!}{%
    \begin{tabular}{cccc}
    \toprule
        \rowcolor{Apricot}Variants                          & \multicolumn{3}{c}{DAVIS-17 Val}                                                      \\ 
        \cline{2-4}
         \rowcolor{Apricot}                                        & \multicolumn{1}{c}{$\mathcal{J}$\&$\mathcal{F} \uparrow$}           & \multicolumn{1}{c}{$\mathcal{J} \uparrow$}             & $\mathcal{F} \uparrow$           \\ 
        \midrule
        KL-divergence                    & \multicolumn{1}{c}{54.65}         & \multicolumn{1}{c}{52.10}          &  57.20       \\ \hline
        CKA                    & \multicolumn{1}{c}{\textbf{72.75}}         & \multicolumn{1}{c}{\textbf{69.90}}          &    \textbf{75.60}     \\ 
        \bottomrule
    \end{tabular}}
    \label{table:distance_metrics}
\end{minipage}%
\hfill%
\begin{minipage}[t]{.56\linewidth}
\centering
\caption{Comparison of different combinations of attention maps ($\mathcal{L}_{att}$) and logit layers ($\mathcal{L}_{inter}+\mathcal{L}_{intra}$) used in the distillation process.}
\resizebox{0.8\textwidth}{!}{%
    \begin{tabular}{cccc}
    \toprule
        \rowcolor{Apricot}Variants                          & \multicolumn{3}{c}{DAVIS-17 Val}                                                      \\ 
        \cline{2-4}
        \rowcolor{Apricot}                                       & \multicolumn{1}{c}{$\mathcal{J}$\&$\mathcal{F} \uparrow$}           & \multicolumn{1}{c}{$\mathcal{J} \uparrow$}             & $\mathcal{F} \uparrow$           \\ 
        \midrule                                
        Att. map 1, logit                    & \multicolumn{1}{c}{ 69.65}         & \multicolumn{1}{c}{66.90}          &    72.40     \\ \hline
        Att. map 2, logit                    & \multicolumn{1}{c}{66.40}         & \multicolumn{1}{c}{63.80}          &     69.00    \\ \hline
        Att. map 3, logit                    & \multicolumn{1}{c}{\textbf{72.75}}         & \multicolumn{1}{c}{\textbf{69.90}}          &    \textbf{75.60}     \\ 
        \bottomrule
    \end{tabular}
    }
    \label{table:setting}
\end{minipage} 
\end{table*}

\paragraph{Loss functions} Recall that we propose to use the CKA score~\cite{kornblith2019similarity} to define the loss for attention distillation in~\cref{eq:att_loss}. We prove that such a selection is effective. Specifically, we compare the CKA with the commonly used KL divergence in implementing the attention distillation loss. We present this comparison in Fig.~\ref{fig:losses}. We observe that the CKA loss is well below the KL loss and clearly shows its convergence. This result also illustrates the anisotropic property of the KL loss in KD. We also investigate the logit distillation loss ($\mathcal{L}_{inter} + \mathcal{L}_{intra}$) in~\cref{eq:logit_loss} and the entire loss ($\mathcal{L}$) in~\cref{eq:dist_loss} in Fig.~\ref{fig:losses}. As shown, our loss functions converge during the KD process. We quantitatively evaluate these distance metrics in terms of the performance of VOS. The results of this study is reported in Table~\ref{table:distance_metrics}, which clearly confirms our choice (i.e., CKA) for the attention loss.

\paragraph{Distillation information} An important question to KD applications is what information should be distilled. In this ablation study, we experiment with different combinations of attention and logit layers used in the distillation process. Specifically, we choose attention maps generated from the GPM of the ID branch from the teacher model (see Fig.~\ref{fig:framework}(a)). Recall that our student network is trained by distillation of the attention map and the logit layer in the 3rd GPM from the teacher network. We report the performance of these combinations in Table~\ref{table:setting}, which clearly confirms the best performance of our student network.

\section{Conclusion}
\label{sec:conclusion}

We propose a novel method for video object segmentation (VOS) with knowledge distillation mechanism. The novelty of our work lies in improving attention learning to adapt with temporal changes in VOS via deformable attention that allows flexible feature locating, and a new knowledge distillation framework that enhances the distillation process via attention transfer. 

We apply these technical innovations to create and train a lightweight VOS network. The network is shown to be adapted to both the spatial and temporal dimensions. We evaluate our method through extensive experiments on several benchmark datasets. Experimental results verify the robustness and efficiency of our method, and show that our method achieves state-of-the-art performance on YouTube-VOS18 and optimal memory usage. 

\section*{Acknowledgements}
This work was partially supported by an internal grant from HKUST (R9429). The authors would like to express their sincere gratitude to the ``Sustainable Smart Campus as a Living Lab'' (SSC) program at HKUST for its invaluable support.

%
%
\bibliographystyle{splncs04}
\bibliography{main}
\end{document}